\documentclass[11pt,a4paper]{article}
\usepackage[hyperref]{acl2018}
\usepackage{times}
\usepackage{latexsym}

\usepackage{url}

\usepackage{graphicx}

\aclfinalcopy % Uncomment this line for the final submission
\title{Improv Chat: Second Response Generation for Chatbot}

\author{Furu Wei \\
  Microsoft Research Asia, Beijing, China \\
  {\tt fuwei@microsoft.com}
}

\date{}

\begin{document}
\maketitle
\begin{abstract}
  Existing research on response generation for chatbot focuses on \textbf{First Response Generation} which aims to teach the chatbot to say the first response (e.g. a sentence) appropriate to the conversation context (e.g. the user's query). In this paper, we introduce a new task \textbf{Second Response Generation}, termed as Improv chat, which aims to teach the chatbot to say the second response after saying the first response with respect the conversation context, so as to lighten the burden on the user to keep the conversation going. Specifically, we propose a general learning based framework and develop a retrieval based system which can generate the second responses with the users' query and the chatbot's first response as input. We present the approach to building the conversation corpus for Improv chat from public forums and social networks, as well as the neural networks based models for response matching and ranking. We  include the preliminary experiments and results in this paper. This work could be further advanced with better deep matching models for retrieval base systems or generative models for generation based systems as well as extensive evaluations in real-life applications.
\end{abstract}

\section{Introduction}

%A chatbot is a computer program or an artificial intelligence which conducts a conversation via auditory or textual methods.
Recent advance in chatbot research and application mainly comes from, (1) the large volume conversation corpus (in the form of query and response pairs) collected from publicly available conversations from online forums and social networks; (2) the neural networks based models trained with the large-scale query-response pairs, which are used as ranking features in retrieval based systems~\cite{DBLP:journals/corr/JiLL14} or end-to-end generative models~\cite{DBLP:journals/corr/VinyalsL15} in generation based systems. 
Usually, it is the users' burden to keep the conversation going, especially when the response from the chat is too short or generic. We tackle this challenge by introducing a new task, termed as Improv chat, which aims to generate a second response after the first response with respect the conversation context. The chatbot can therefor say more so as to continue the conversation.

\begin{figure}
	\centering
	\includegraphics[width=0.9\linewidth]{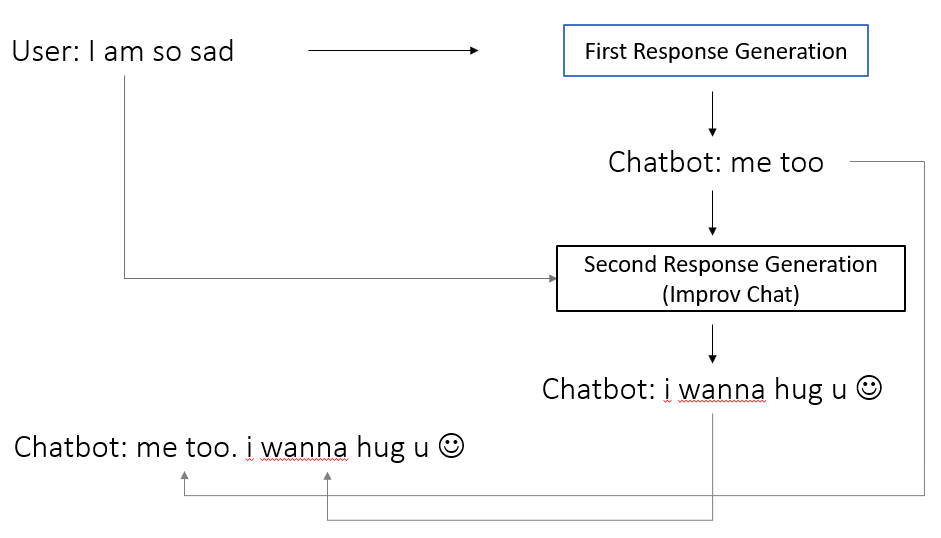}
	\caption{Example of Improv chat}
	\label{fig:improv_example}
\end{figure}

Figure~\ref{fig:improv_example} shows the overview and framework of a chatbot system with Improv chat capability. Given the user's input ``i am sad'', the first response generation system, which can be either a retrieval based system or generation based system, may return a shot response as ``me too''. The response is too short and we will trigger the second response generation system (i.e. Improv chat) and return a improv response as ``i wanna hug u''. The system will then combine the results from both the first response generation and the second response generation and return the final response as ``me too. i wanna hug u'' to the user, which is more informative and emotional and will thus improve the user engagement and user experience of the chatbot.

\begin{figure*}
	\centering
	\includegraphics[width=0.7\linewidth]{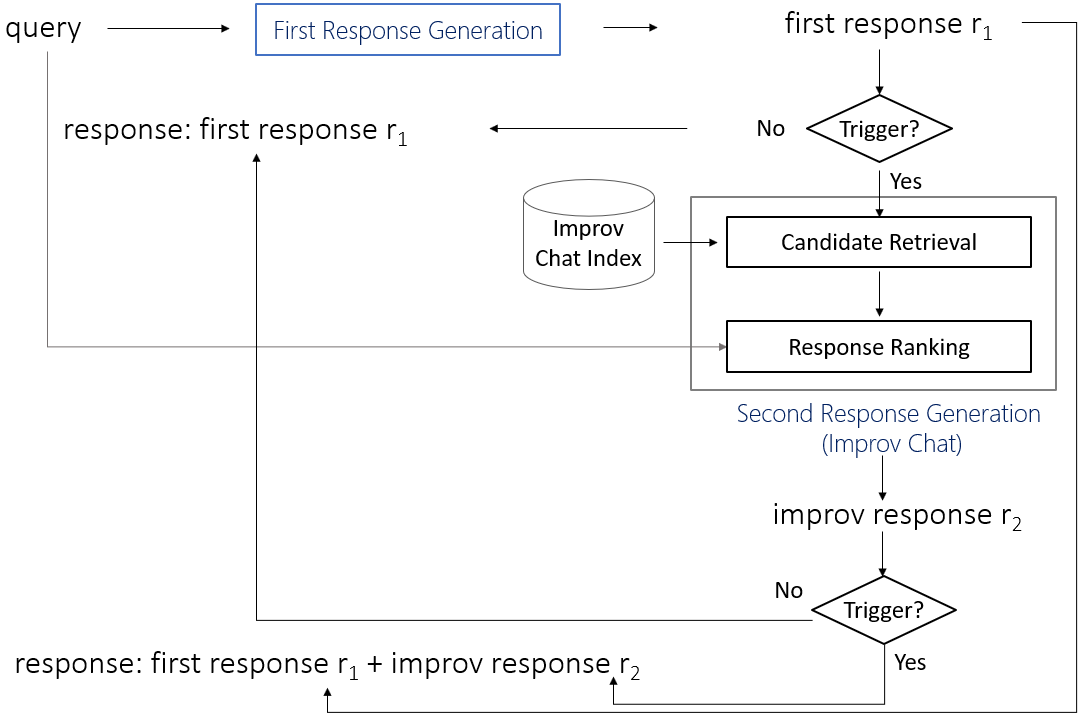}
	\caption{Framework of Improv chat}
	\label{fig:improv_framework}
\end{figure*}

\section{Improv Chat}
%
%Similar to the first response generation system, 

The second response generation system, namely the Improv chat, can be implemented with retrieval based approach or generation based approach. In this paper, we implement the Improv chat with retrieval based method. Figure~\ref{fig:improv_framework} shows the general framework of the learning based Improv chat system. Given the users query \textit{q}, we will first use the first response generation system to generate the first response $r_1$. If the Improv chat mechanism is triggered, we will use $r_1$ to retrieve the candidate Improv responses from the Improv chat index which is built offline. Then we use the query $q$ as the context query to rank the candidate Improv responses to make sure that the top ranked Improv responses are relevant and appropriate to the conversation context. We then select the Improv response $r_2$ (with randomness) from top ranked and relevant Improv candidates filtered by response ranking. If we aim to trigger the Improv response $r_2$, the chatbot will return $r_1 + r_2$ as the final response to the user. The two key parts are how to collect the conversation corpus for Improv chat and how to design the response ranking algorithm. We will elaborate the details in the following sections and we will also introduce the triggering logic for Improv chat.

%\begin{figure}
%	\includegraphics[width=\linewidth]{improv0.png}
%	\caption{Examples of Improv chat}
%	\label{fig:improv_example}
%\end{figure}

\begin{figure*}
	\centering
	\includegraphics[width=0.7\linewidth]{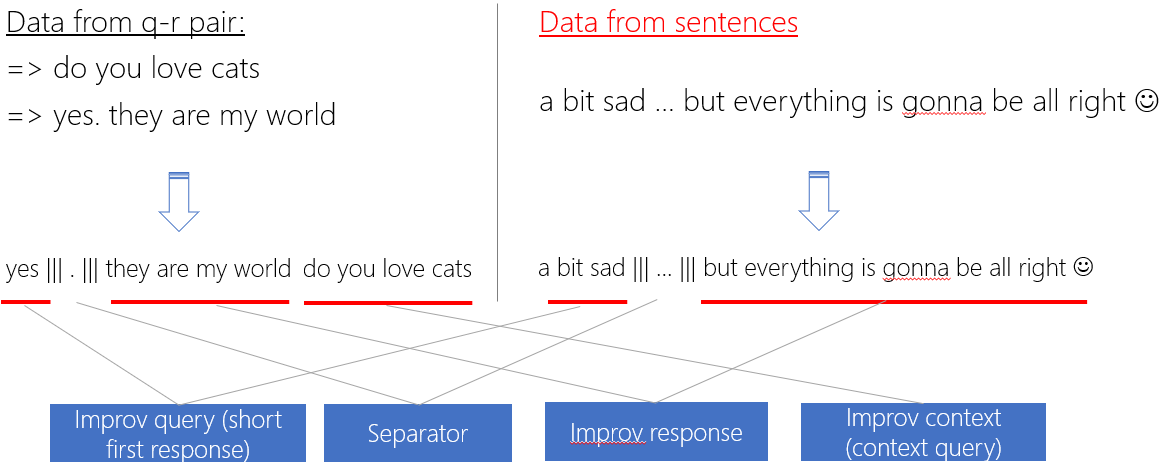}
	\caption{Data collection for Improv chat}
	\label{fig:improv_data}
\end{figure*}

\subsection{Data}

In general, the conversation corpus for the Improv chat engine is in the format of $<$\textit{short response, improv response, context query}$>$, where \textit{improv response} is the second response which could be appended to the \textit{short response} in the context of \textit{context query}. As shown in Figure~\ref{fig:improv_data}, we propose and implement two methods to collect the conversation corpus to build the retrieval based Improv chat system. First, it could be extracted from the query-response pairs collected from online forums and social networks which are used build the first response generation system. As shown in the left hand side of Figure~\ref{fig:improv_data}, given the query ``do you like cats'' and the response ``yes. they are my world'', if the response can be spitted into two parts with punctuations for sentence segmentation (e.g. ``.', ``'?'', ``...'' etc.) and the first part is a short sentence (e.g. less than 5 words), we can then extract ``yes'' as the \textit{short response}, ``they are my world'' as \textit{improv response} and ``do you like cats'' as \textit{context query}. Second, we can also extract the conversation corpus from general chat sentences, as shown in the right hand side of Figure~\ref{fig:improv_data}. Given a chat sentence ``a bit sad ... but everything is gonna all right :)'', we can also split it into two parts using the same method and logic as in extracting Improv chat data from query-response pairs mentioned previously. We can extract ``a bit sad'' as the \textit{short response} and ``but everything is gonna all right :)'', and there will be no \textit{context query}. We can collect a large-scale conversation corpus to build the Improv chat index. We collect more than 74 million pairs in our experiments to build the prototype system. We use Lucene~\footnote{https://lucene.apache.org/} to build the Imprvo chat index with the \textit{short response} field as the searchable field.

\subsection{Ranker}

We use the user's query $q$ to obtain the first response $r_1$ from the first response generation system. In order to get the Improv response for $r_1$ in the conversation context of $q$, we first use the short response $r_1$ and the default Lucene ranking mechanism to retrieve top N most similar short responses $\{q^*\}$ in the Improv chat index. Then we collect the corresponding Improv responses $\{r_2^*\}$ to  $\{q^*\}$. We then use a ranker to rank $\{r_2^*\}$ so that the top ranked candidates are relevant and appropriate with respect to the conversation context $q$. Generally, the Improv response ranking is a very common task which is similar to document ranking in information retrieval. We can reuse the ranking algorithms used in web search. For simplicity, in this paper, we treat it as a classification problem. Given $<q, r_2^+ \in \{r_2^*\}>$, we classify the pair to be relevant (1) or not (0), and use the scores from the classifier as the ranking scores. We implement the classifier with libsvm~\footnote{https://www.csie.ntu.edu.tw/~cjlin/libsvm/}, with the following ranking features.

\begin{itemize}
	\item Translation model: we train a translation model~\cite{DBLP:journals/corr/JiLL14} with the query-response pairs used in the first response generation, which are also collected from public forums and social networks. The scores calculated from $q$ and $r_2^+$ by the translation model are used as features.
	\item Deep matching model: we train a neural networks based matching model using similar networks structure as in~\citet{smarreply17}. The matching score from the neural networks with $q$ and $r_2^+$ is used as a feature.
	\item Language model: we train as neural language model with the sentences from the conversation corpus, and use the language model score calculated on $r_2^+$ as a feature.
\end{itemize}

We use 5,000 manually labeled pairs to train the response ranking model (i.e. the classifier) for Improv chat. The precision of relevant class is 0.74 at the recall of 0.75.

\subsection{Triggering}

We use the following logic to design the triggering mechanism of Improv chat.

\begin{itemize}
	\item The Improv chat will only triggered for short responses (e.g. less than 5 words) from the first response generation system. 
	\item It will be triggered more for passive users. 
	\item There will be randomness in triggering Improv chat. While keeping the conversation going is important, the system should not interrupt a user if he/she have something to talk to the chatbot.
	
\end{itemize}

\section{Conclusion}

We introduce a new task, namely Improv chat, for response generation for chatbot in this paper. We propose a general data-driven framework to model Improv chat and develop a retrieval based system to generate the second response with users' query as context and the result from first sentence generation as input. The first response is used to retrieval candidate second responses from the Improv chat index which is built from public forums and social networks. We then use the users' query as the context to rank the candidate second responses to make the top ranked responses relevant and appropriate. The response from the chatbot will be concatenated by the results from both the first response generation and second response generation (i.e. Improv chat) which will more informative and emotional. This work could be further advanced with better deep matching models for retrieval based systems as well as end-to-end generative models for Improv chat with sequence-to-sequence learning.

\section*{Acknowledgments}
We thank Xiaopeng Wu from Microsoft Bing Search team, Pengjie Ren from Shandong University, Shaohan Huang, Lei Cui, and Nan Yang from Microsoft Research Asia for valuable discussions and data preparation. 

%\bibliography{acl2018}
\bibliography{acl2018}
\bibliographystyle{acl_natbib}

\end{document}